\documentclass{article}

\usepackage{graphicx}
\usepackage{float}
\usepackage{multirow}
\usepackage{amsmath}
\usepackage[preprint]{neurips_2024}

\usepackage[utf8]{inputenc} 
\usepackage[T1]{fontenc}    
\usepackage[colorlinks=true, linkcolor=blue]{hyperref}       
\usepackage{url}            
\usepackage{booktabs}       
\usepackage{amsfonts}       
\usepackage{nicefrac}       
\usepackage{microtype}      
\usepackage{xcolor}         
\usepackage{cleveref}
\usepackage[colorlinks=true, linkcolor=blue]{hyperref}

\title{SynRailObs: A Synthetic Dataset for Obstacle Detection in Railway Scenarios}

\author{
  Qiushi Guo \\
  Independent Researcher \\
  Chengdu, China \\
  \texttt{guoqiushi910@gmail.com} 
  \And
  Jason Rambach \\
  DFKI \\
  Kaiserslautern, Germany \\
  \texttt{Jason\_Raphael.Rambach@dfki.de} \\
}

\begin{document}

\maketitle

\begin{figure}[H]
    \centering
    \includegraphics[width=0.9\linewidth]{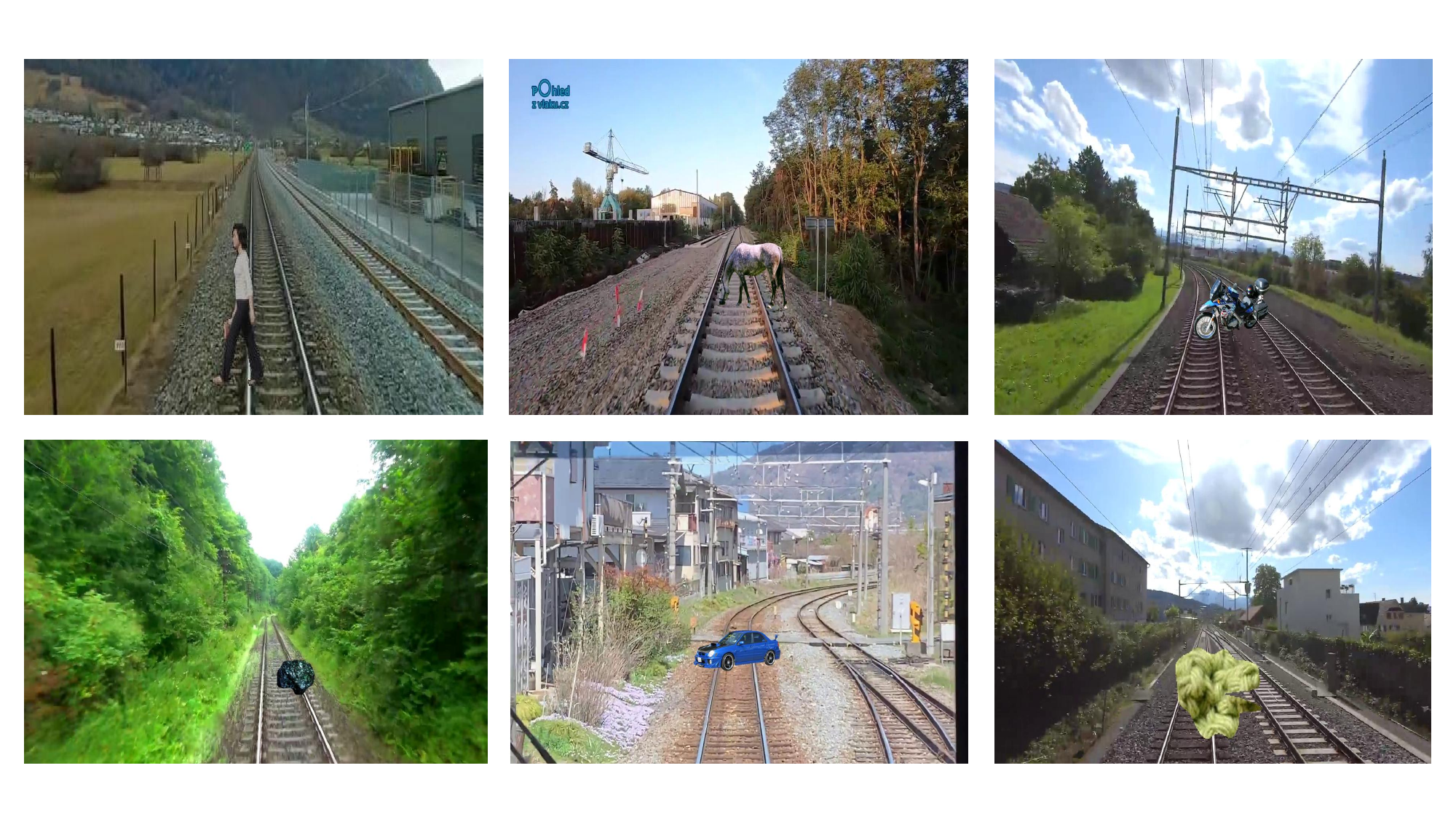}
    \caption{Sample images in SynRailObs. From top to bottom, left to right are
    persons, animals, motorcycle, rocks, vehicles and random polygons.}
    \label{fig:enter-label}
\end{figure}
\begin{abstract}
Detecting potential obstacles in railway environments is critical for preventing serious accidents. Identifying a broad range of obstacle categories under complex conditions requires large-scale datasets with precisely annotated, high-quality images. However, existing publicly available datasets fail to meet these requirements, thereby hindering progress in railway safety research. To address this gap, we introduce SynRailObs, a high-fidelity synthetic dataset designed to represent a diverse range of weather conditions and geographical features. Furthermore, diffusion models are employed to generate rare and difficult-to-capture obstacles that are typically challenging to obtain in real-world scenarios. To evaluate the effectiveness of SynRailObs, we perform experiments in real-world railway environments, testing on both ballasted and ballastless tracks across various weather conditions. The results demonstrate that SynRailObs holds substantial potential for advancing obstacle detection in railway safety applications. Models trained on this dataset show consistent performance across different distances and environmental conditions. Moreover, the model trained on SynRailObs exhibits zero-shot capabilities, which are essential for applications in security-sensitive domains. The data is available in \href{https://www.kaggle.com/datasets/qiushi910/synrailobs}{\textcolor{blue}{https://www.kaggle.com/datasets/qiushi910/synrailobs}}.

\end{abstract}

\section{Introduction}
With the rapid development of high-speed railways, the demand for enhanced railway security has been continuously increasing. Traditional approaches face significant challenges in addressing the complexities of railway environments, particularly due to variable weather and lighting conditions. Deep learning has achieved remarkable success across a wide range of domains. Deep learning-based computer vision techniques have been successfully applied in areas such as remote sensing, fraud detection, and medical diagnosis, demonstrating significant potential for industrial applications.

Deep learning techniques have been increasingly applied to railway security scenarios, with approaches based on classification, object detection, and segmentation being deployed in real-world settings. Previous studies have shown that deep learning-based models yield satisfactory results when the training data shares a similar distribution with the input data in practical applications. However, their performance drops significantly when there is a distribution shift. The task of obstacle detection is particularly affected by this issue. In contrast to the vast number of real-world scenarios, the available training data is limited both in quantity and diversity.

The challenges in railway obstacle detection arise from several factors: First, distribution gap. The complexity of the environment. Railways can be located in urban areas, mountainous terrains, or near lakes, pass through tunnels, and be surrounded by various natural elements such as branches and flowers. Additionally, weather conditions are highly variable, further complicating the data distribution. Features associated with snow, rain, and fog are difficult to capture with a limited amount of training data. Second is the Zero-shot problem. Potential obstacles are highly unpredictable and may not be represented in the training set. Existing public railway-related datasets either lack the diversity of obstacles and environmental conditions or have insufficient quantities of images and corresponding annotations. As a result, these datasets do not meet the requirements of current obstacle detection tasks and hinder the advancement of railway security in both academia and industry.

To address the aforementioned challenges, we propose the SynRailObs dataset, which aims to mitigate these issues to some extent. Given the difficulty of collecting realistic obstacle images from practical railway environments, simulation presents the only viable solution. Previous approaches, such as using game simulators like Train Sim World, have been employed to simulate such scenarios. However, a significant gap remains between images generated by game engines and those that reflect real-world conditions. The synthetic images can be decomposed into three components: background, foreground, and harmony. Our pipeline is designed to address both the distribution gap and Zero-shot issues. For the background, we scrape videos of trains traveling on railway tracks across different countries and under various weather conditions. For the foreground, we gather potential obstacles from both public datasets and generative models. The combinations and permutations of these factors allow the data distribution to closely resemble that of the real world.

To address the Zero-shot problem, we randomly generate polygons to represent contours of potential unseen obstacles and fill them with random textures from the DTD dataset. These objects are then extracted from the generated images using SAM (Segment Anything Model) and pasted onto background images, taking into account railway track features and affine transformations. At this stage, the generated images exhibit inconsistencies between the foreground elements and the background. To enhance realism, we apply an image harmony technique to improve the visual coherence of the composite images.

The entire process is fully automated and free of manual annotation, ensuring both efficiency and error-free generation
Comprehensive experiments have been conducted to validate the effectiveness of SynRailObs. Our dataset demonstrates its robustness in real-world railway obstacle detection scenarios across various settings. The model's performance remains relatively stable across different distances, and models trained on SynRailObs are capable of detecting obstacles even under adverse weather conditions. Additionally, SynRailObs effectively addresses the zero-shot problem.  
Our contributions can be summarized as follow:
\begin{itemize}
    \item We propose a highly realistic synthetic dataset named SynRailObs, designed to address the data shortage issue in railway obstacle detection scenarios. SynRailObs includes a wide range of potential obstacles, such as pedestrians, rocks, animals, vehicles, and more. The background images in SynRailObs encompass various countries and feature diverse, complex weather conditions.
    \item We propose a pipeline that leverages Vision-Language Models (VLMs) such as Stable Diffusion, along with the SAM (Segment Anything Model) and image harmony techniques, to generate highly realistic synthetic images. Models trained on the synthetic images generated by this pipeline demonstrate performance comparable to those trained on real-world images.
    \item We conduct experiments in practical scenarios to validate the effectiveness of SynRailObs. The results demonstrate that our dataset is capable of handling railway obstacle detection tasks in real-world conditions, including complex weather scenarios and zero-shot cases.
\end{itemize}
\section{Related Work}

\subsection{Railway Obstacle Detection}
Railway obstacle detection can generally be classified into three categories: vision-based, radar-based, and vibration-based approaches. Radar-based methods, which include Light Detection and Ranging (LiDAR) and millimeter-wave (mmWave) radar, offer relatively stable performance under diverse weather conditions and are less affected by lighting changes. These methods also provide a considerable detection range, often extending to hundreds of meters. \cite{dias2024lidar} propose a LiDAR-based approach capable of detecting obstacles at distances up to 500 meters. \cite{qu2023research} introduce an obstacle detection method based on LiDAR, which captures and processes rich three-dimensional (3D) information and depth data from the railway scene. \cite{tang2023moving} integrate millimeter-wave radar with the DBSCAN clustering algorithm and velocity filtering techniques to achieve accurate and reliable detection of dynamic targets along the train’s path. \cite{zhao2023millimeter} propose a new method for detecting obstacles in front of trains using millimeter-wave radar in highly dynamic railway environments. This method combines radar projection segmentation with Kalman filtering, enabling real-time and high-accuracy obstacle detection. Despite the advantages of radar-based approaches, their limitations should not be overlooked. The deployment of either LiDAR or mmWave radar is more expensive compared to cameras, which hinders the widespread adoption of radar technology in railway scenarios.

As a result, vision-based methods have gained increasing attention in recent years. \cite{brucker2023local} propose using a shallow neural network to learn railway segmentation from typical railway images. The network’s limited receptive field prevents overconfident predictions and allows it to focus on the locally distinct and repetitive patterns typical of the railway environment. \cite{zhang2024improving} introduce a small-sample object detection algorithm, YOLOv5-RTO, which employs the EVC (Explicit Visual Center) attention mechanism and a lightweight up-sampling operator, CARAFE (Content-Aware ReAssembly of Features), to enhance YOLOv5’s performance. \cite{guo2024universal} propose an optical-flow-guided segmentation method for detecting obstacles in railway scenarios.
\subsection{Synthetic Dataset}
Training deep neural models, particularly Vision-Language Models (VLMs), typically requires large volumes of data. While real-world data is often the ideal choice, access to task-specific datasets is frequently limited due to privacy concerns and the high costs associated with data annotation. In such cases, synthetic data offers a viable alternative to address these challenges.

One intuitive approach for generating synthetic data is image composition. For instance, \cite{composite} proposed a task-aware method that generates synthetic data through image composition. The "copy-paste" technique has also demonstrated effectiveness in creating realistic synthetic datasets. Another approach involves utilizing game engines to generate images. \cite{GTA} introduced a method for rapidly generating pixel-accurate semantic label maps for images extracted from modern video games, specifically GTA5. This dataset has been widely used to validate cross-domain tasks. Similarly, \cite{trainsim} employed a synthetic dataset for railway scenarios, facilitating the simulation of related tasks.

Generative models have also been leveraged to produce realistic data from scratch. For example, \cite{melzi2023gandiffface} proposed a novel framework combining Generative Adversarial Networks (GANs) and Diffusion Models to generate synthetic datasets for face recognition, overcoming the limitations of existing synthetic datasets. Additionally, \cite{david_beniaguev_2022_SFHQ} introduced a high-resolution synthetic dataset based on StyleGAN and Stable Diffusion, which offers rich annotations and is particularly valuable for training tasks that require detailed, high-quality data.

\subsection{Railway dataset}
Previous datasets in railway scenarios have focused on rail surface defect detection, rail segmentation, foreign object detection, and related tasks. \cite{railway3d} propose a diverse point cloud semantic segmentation (PCSS) dataset specifically designed for railway environments. \cite{trainsim} introduce a simulation framework, TrainSim, capable of generating multiple types of datasets. \cite{raildb} present a real-world railway dataset, Rail-DB, which includes 7,432 pairs of images and corresponding annotations. These images are collected under varying lighting conditions, road structures, and perspectives. \cite{railsem19} consists of 8,500 annotated short sequences captured from the ego-perspective of trains, including over 1,000 examples of railway crossings and 1,200 tram scenes.

However, to the best of our knowledge, no obstacle detection datasets have been proposed specifically for railway scenarios, let alone datasets that encompass multiple types of obstacles and cover complex environments.
\section{SynRailObs}
\begin{figure}[t]
    \centering
    \includegraphics[width=0.9\linewidth]{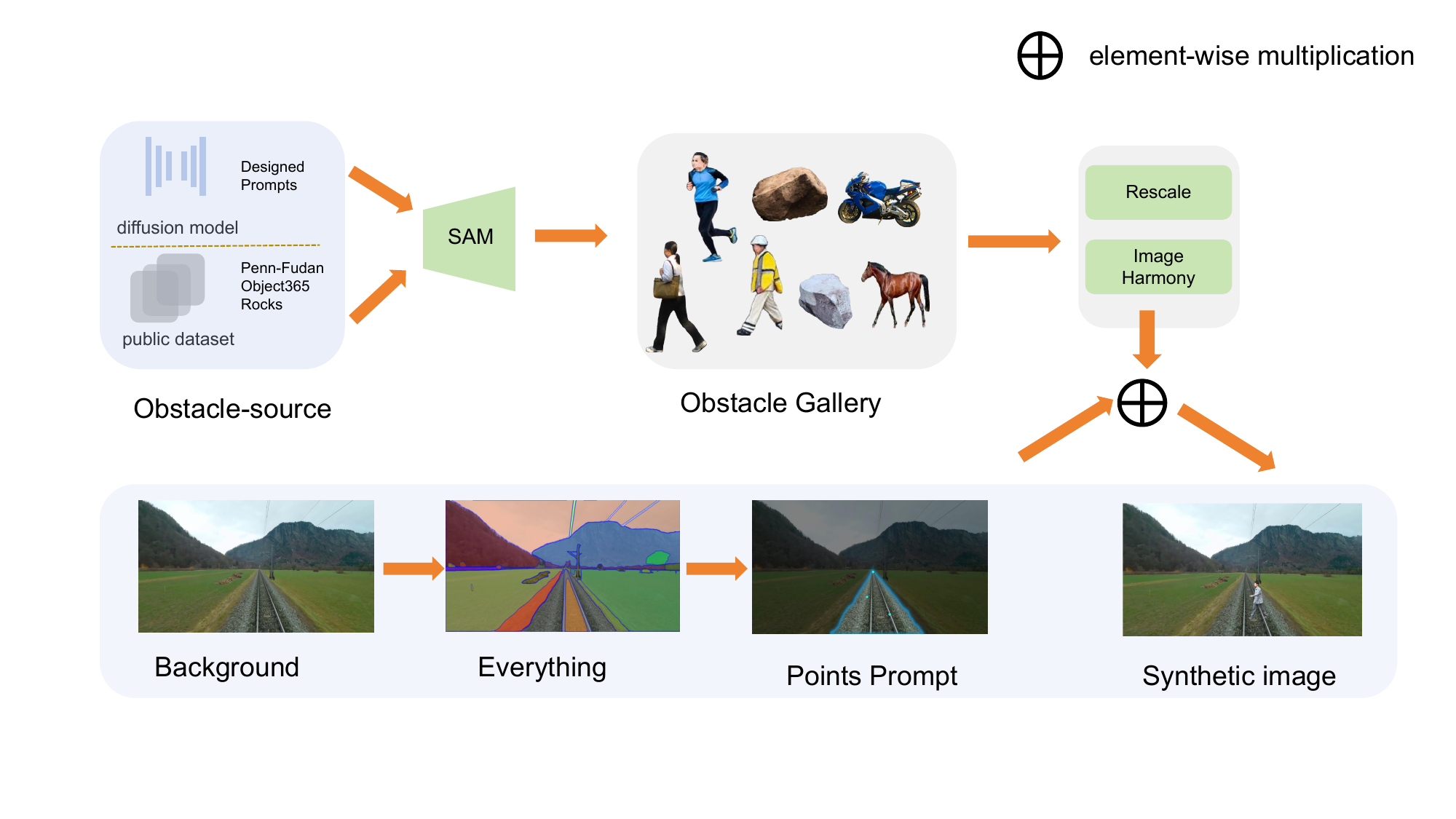}
    \caption{Workflow of synthetic image generation. In obstacle brach, Potential obstacles are extracted from stable-diffusion generated images and public datasets guided by SAM. The extracted obstacles form an obstacle gallery; In background branch, railway area are determined by SAM leveraging point prompts. The obstacles are pasted on railway area after rescale and harmonization. }
    \label{fig:enter-label}
\end{figure}
In this part, we first introduce the materials of SynRailObs, namely background images and foreground objects. Background images are frames extracted from web videos and foreground objects are segmented from public auxiliary datasets and   Then we will show the workflow of image composition. After this step, image harmony is introduced to retain the consistency of background and foreground. The details of each step are demonstrated in corresponding parts. 
\subsection{Background Images}
The background images are derived from videos of trains in motion. These videos were sourced from the internet under a non-commercial usage agreement. The geographical coverage spans a wide range, from Europe to Asia, and the weather conditions vary, ensuring a diverse set of image features. The original images do not contain any obstacle-related incidents, and no obstacles are present within the railway zones. To streamline the image generation process, images with potential obstacles are excluded. As a result, only images devoid of people, vehicles, and bicycles are used as background images. Annotations for synthetic obstacles can be generated automatically.
\begin{table}[htb]
    \centering
    \begin{tabular}{c|c|c|c}
    \hline
       Country  & weather&volume&resolution\\
       \hline
       \centering China, Vietnam, Italy, Switzerland, Spain  &Sunny, rainy, snowy&10000&1280$\times$720\\
       \hline
       
    \end{tabular}
    \vspace{0.5cm}
    \caption{Description of background images}
    \label{tab:my_label}
\end{table}
\subsection{Stable-Diffusion Data}
Current public datasets can not fully meet our requirements in several aspects.
For instance, the pedestrian in public datasets are frequent occluded, which lead to
incomplete and counterintuitive generated synthetic images; public pedestrians lack of diversity of pose, which is crucial in railway scenarios. Besides, certain sub-categories, like railway maintenance workers or security patrol personnel are seldom
in public datasets. The similar dilemmas occur in other obstacles, which hinder the 
improvement of models.
To this end, Stable-diffusion are leveraged to generate to enrich desired target obstacles. We design prompts in following format:
\begin{table}[htb]
    \centering
    \begin{tabular}{cc}
    \toprule
     categories    &  prompts\\
     \midrule
     Pedestrians    &``In \{weather\} days, \{num\} pedestrians \{motion\} through {places}''\\
     \midrule
     Rocks & ``\{num\} rocks in \{distance\} meters far away in the \{pos\} on \{place\}''\\
     \midrule
     Animals &`` \{num\} of \{species\} \{motion\} in \{places\} in \{weather\} at\{time\}''\\
     \midrule
     vehicles & ``\{num\} of \{vehicle\_type\} \{motion\} in \{place\} in \{weather\} days at \{time\}''\\
     \midrule
     \bottomrule
    \end{tabular}
    \vspace{0.5cm}
    \caption{Prompts template in generating obstacle images.}
    \label{tab:my_label}
\end{table}
The placeholder like weather, num, distance, motion, etc. enlarge the diversity of the generated objects.
We design a 
The generated objects lose part details compared to realistic counterparts, however, the almost zero cost and rich content ensure the coverage of the distribution. 
\subsection{Auxiliary Datasets}
\paragraph{Object365} \cite{objects365}is a large-scale dataset, designed to spur object detection research with a focus on diverse objects in the Wild. It has 365 categories with 2 million images and over 30 million bounding boxes. 
The wide variety of categories in object365 cover the frequent obstacles in Railway scenarios. Images with Person, Motocycle, animals, cars, etc., can be used directly serving as obstacles to generate images. The corresponding bounding box can be fed into SAM as prompt to extract mask of target objects without extra annotations.
\paragraph{PennFudan}\cite{penn-fudan} is an image dataset containing images for pedestrian detection/segmentation. The images are taken from scenes around campus and urban street. The heights of labeled pedestrians in this database fall into [180,390] pixels. All labeled pedestrians are straight up. There are 170 images with 345 labeled pedestrians, among which 96 images are taken from around University of Pennsylvania, and other 74 are taken from around Fudan University. The pedestrians in this dataset are quite similar to our railway scenarios, regarding to the pose and contour completeness.
\paragraph{Describable Textures Dataset (DTD)}\cite{dtd} is an evolving collection of textural images in the wild, annotated with a series of human-centric attributes, inspired by the perceptual properties of textures. DTD consisting of 5640 images with 47 items, 120 images for each category. Image sizes range between 300x300 and 640x640. 
We use DTD to increase the generalization ability of models trained on our dataset. The obstacles filled with different textures in DTD with both fixed and random contour
can improve the performance in zero-shot scenarios.
\paragraph{Rock Images} \cite{rockclassification2022} is a dataset of rocks with 53 folders of various rocks, each folder contain around 45 images. The images are collected from Internet. The images can not be leveraged directly due to the lack of annotations. The rocks are extracted guided by SAM and yolo
serving as falling rocks and mudslides in our railway scenarios. 

\subsection{Image composition}
The composition workflow is demonstrated as. Target images from different sources(stable-diffusion, public datasets) are randomly sampled and then fed into
SAM with corresponding bounding boxes. The bounding box can be obtained by pretrained-yolo if not provided in advance. The extracted target objects form objects gallery.
As for the base image branch, the background image is fed into a rail segmentation model to locate the rail area. Since the pretrained rail segmentation result is not reliable compared to the SAM. We sample points and bounding box based from its logits
serving as prompts to obtain accurate results from SAM. To determine where to paste targets, points inside railway area are sampled as pasted upper-left coordinate. Considering the perspective transformation, the pasted target size should be adjusted along with the pasted coordinate. 
\begin{equation}
    \hat{W} = \alpha\cdot w+\beta
\end{equation}
Since the railway are not always straight, $\alpha$ ranges from 0.4 to 0.6, $\beta$ ranges from 30-45. The pasted process can be formulated as follow:
\begin{equation}
    I_{syn} = I_{T}\otimes M +\mathbf{B}\otimes(1-M)
\end{equation}
where $I_{T}$ is target image, $\mathbf{B}$ is background image, and $M$ is mask extracted from SAM. 

The generated images in this stage show inconsistency between pasted obstacles and background due to the distribution gap between illumination and resolution.  Image harmony technique is introduced to alleviate this issue. Given a foreground image $F$ with corresponding mask $M$ and a background image $B$, the task of image harmonization
is to design a function $\psi(\cdot)$, to create a natural image $I$:
\begin{equation}
    I= \psi(M,F)+(1-M)\mathbf{B}
\end{equation}
In this work, we adopt an encoder-based approach to harmonize the generated images. The method proposed by \cite{harmonizer} is utilized to achieve this task. The harmonized images demonstrate high or at least comparable confidence scores, and the class activation map (CAM) heatmap further validates the effectiveness of employing harmony tools.


\begin{figure}[t]
    \centering
    \includegraphics[width=0.95\linewidth]{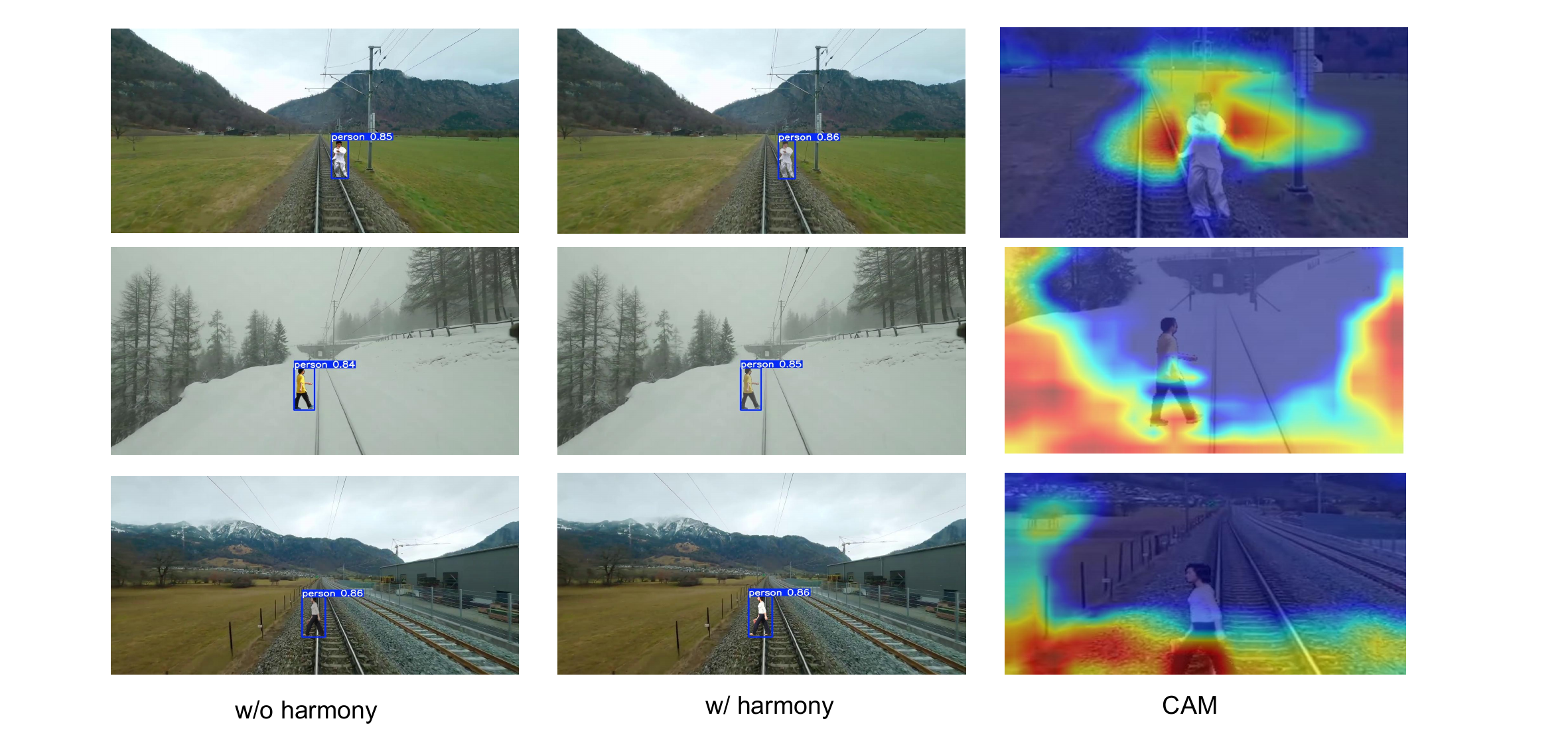}
    \caption{The first and second columns represent the prediction confidences of the YOLO model pretrained on Objects365 for unharmonized and harmonized images, respectively.The third column shows the Grad-CAM visualizations of the corresponding regions.  }
    \label{cam}
\end{figure}
\section{experiments}
\begin{figure}[htb]
    \centering
    \includegraphics[width=0.95\linewidth]{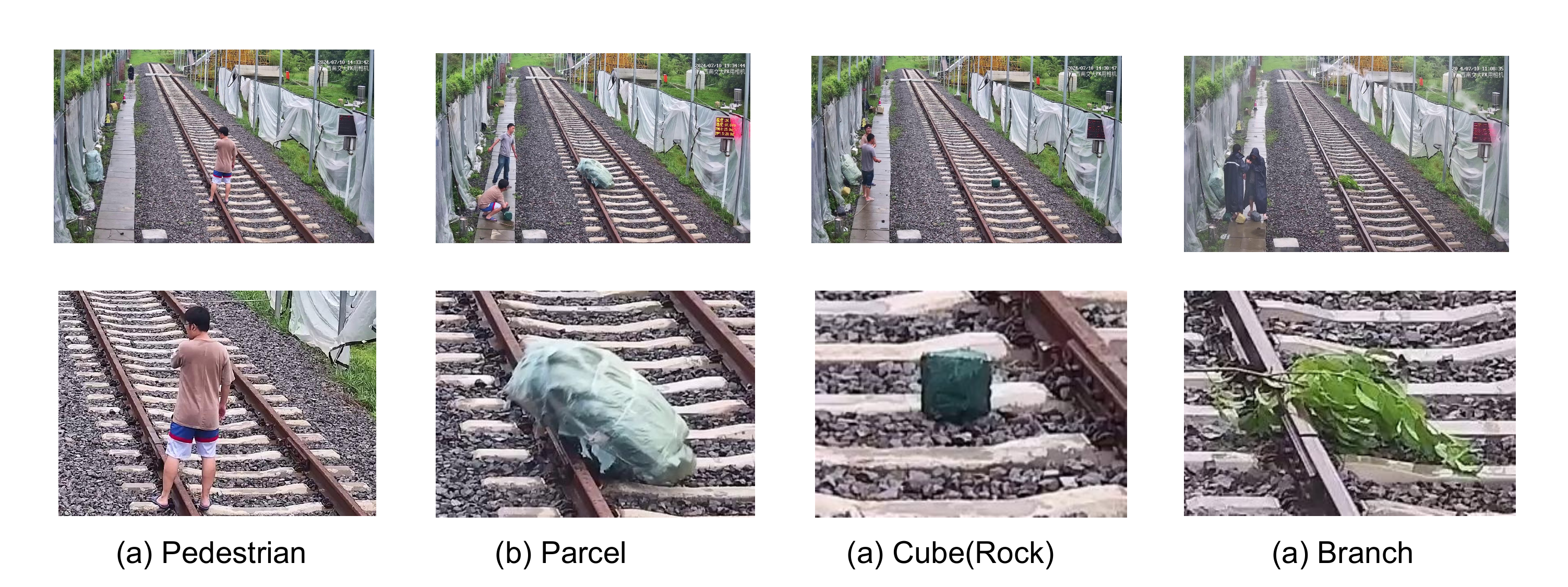}
    \caption{Samples of test images captured in our experimental site. }
    \label{fig:enter-label}
\end{figure}
\subsection{Configuration}
Pytorch is used as our framework. A RTX 3090Ti with memory 24GB and I7-14700KF are utilized to train our models. Batchsize is set to 32, Adam is used as the optimizer. Initiatial learning rate is 0.001. Epoches are set to 50, and the input size of image is 640. 
\subsection{Test dataset}
To verify the effectiveness and usability of proposed SynRailObs, we also prepare a test dataset in practical scenario. The images in test dataset are collected in our experimental site in Chengdu, China. The experiment sites are 80 meters long, and camera is set 4 meters high with resolution 1920x1080.
The categories of obstacles include pedestrians, rocks, branches, parcels and steer board. Besides, ballastless track and ballasted track are both considered to cover various scenarios. Experimental sites are equipped with rain and smog simulator which can simulate complex wether conditions. We collect videos 
under each
\subsection{Metric}
For image-level evaluation, mAP is set as our main metric, which can be calculated as follow:
$$\text{mAP}=\frac{1}{c}\sum^{C}_{i=1}AP_{i}$$
$$\text{AP} = \int_0^1 \text{Precision}(r) \, d\text{Recall}$$
For event-based evaluation, Accuracy is used as the metric. 
$$Accuracy =  \frac{TP+TN}{TP+TN+FP+FN}$$
Each test event is a 15 seconds video, we extract 10 frames per second. The final results are based on the majority vote of all predictions from each frame.  
\subsection{General Results}
\cref{genral result} presents the overall performance of models trained on SynRailObs in both ballastless and ballasted track scenarios. For the detection task, all models achieve mean Average Precision (mAP) values exceeding 50\%. In the event-based task, classification accuracies surpass 90\%, demonstrating the reliability and effectiveness of our dataset.

\begin{table}[h]
\centering
\begin{tabular}{l c | c c c | c c c}
\toprule
\multirow{2}{*}{Models} & \multirow{2}{*}{Paras (M)} & \multicolumn{3}{c|}{ballastless track} & \multicolumn{3}{c}{ballasted track} \\
&  & mAP50 & mAP50:95 & Acc & mAP50 & mAP50:95 & Acc \\
\midrule
YOLOV5-s\cite{yolov5}       &9.1&57.3  & 53.2     & 93.7  & 59.4  & 57.7  & 95.1 \\
YOLOV5-m\cite{yolov5}       &25.1& 60.1 & 54.7  & 95.2     & 61.7  & 59.3& 95.8 \\
YOLOV5-l\cite{yolov5}       &53.2& 63.4 & 58.1  & 96.7     & 63.7   & 61.4  &94.7    \\
nanodet\cite{nanodet}      &2.44& 56.2 & 53.7  & 92.7     & 57.7   & 53.5  &92.7    \\
Faster-RCNN\cite{faster-rcnn}      &44.3& 57.7 & 52.1  & 92.8     & 58.7   & 51.4  &93.7    \\
RE-DETR\cite{detrs}      &41.3& 65.3 & 61.7  & 97.0     & 66.7   & 60.4  &97.2    \\

\bottomrule
\end{tabular}
\vspace{0.5cm}
\caption{Comparison of models on ballastless track and ballasted track}
\label{genral result}
\end{table}

\subsection{Across Environment}
\cref{environment} presents the results under various external conditions: Normal, Rainy, Foggy, and Dark. The models trained on SynRailObs exhibit stable and robust performance under Normal and Rainy conditions, achieving mAP@50 values ranging from 0.497 to 0.637—indicating effective obstacle detection in typical railway scenarios. However, performance degrades significantly under Foggy and Dark conditions, with mAP@50 values falling below 0.5, which is unacceptable for safety-critical applications. A likely explanation is the lack of sufficient training data for these adverse conditions, as such scenarios are underrepresented in our background image collection.

\begin{table}[htbp]
  \centering
  \begin{tabular}{ccccc}
    \toprule
 &Normal&Rainy&Foggy&Dark \\
\midrule
 YOLOV5-s&58.7  &52.4&38.9&31.8 \\
  nanodet&59.2  &51.1&36.1&30.7 \\ 
   Faster-RCNN&59.1&49.7&40.9  &29.1\\ 
   RE-DETR&63.7&53.9&0.468  &34.1\\ 
\bottomrule
  \end{tabular}
  \vspace{0.5cm}
    \caption{Results across different environmental settings}
  \label{environment}
\end{table}
\subsection{Across Distance}
\cref{distance} shows the detection performance at varying distances between obstacles and the camera in the railway environment. The mAP decreases progressively as distance increases—from over or around 0.7 at 0–15 meters to approximately 0.4 at around 75 meters. At our experimental site, obstacles located 70 meters away appear smaller than 20 pixels, making detection particularly challenging. Moreover, images containing such distant (tiny) obstacles constitute only a small fraction of the dataset, leading to an imbalanced distribution. In summary, models trained on SynRailObs demonstrate robust and reliable performance within the 0–45 meter range. Additional efforts are required to improve detection accuracy at greater distances.

\begin{table}[h]
\centering
\begin{tabular}{cccccc}
\toprule
 &0-15m&15-30m&30-45m&45-60m&60-75m\\
\midrule
 YOLOV5-m&72.4&68.2&61.7&51.7&43.7   \\ 
  nanodet&71.7&69.5&62.1&52.3&42.8 \\ 
   Faster-RCNN&69.3&67.7&62.4&54.8&41.7  \\ 
   RE-DETR&74.2&70.7&64.7&59.4&48.1\\
\bottomrule
\end{tabular}
\vspace{0.5cm}
\caption{Result across distances of obstacles}
\label{distance}
\end{table}

\subsection{Zero-shot Obstacles Detection}
\cref{zero-shot} illustrates the zero-shot generalization ability of SynRailObs. Although certain potential obstacles are not present in the dataset, many of them can still be detected under the label unseen obstacles, which is also used for randomly generated polygonal synthetic objects. Some instances, such as cubes and parcels, are partially detected and misclassified as rocks. However, branches are often missed due to their complex shapes and textures, which pose greater challenges for the model.
\begin{table}[h]
\centering
\begin{tabular}{ccccccc}
\toprule
 &Steel Board&Parcel&Cube(white)&Cube(green) &Branch  \\
\midrule
 YOLOV5-m& 63.1 &65.2&59.4&58.3 &45.1  \\ 
  nanodet&62.7  &63.1&62.1&61.8 &43.7 \\ 
   Faster-RCNN&60.9&64.7&60.3&57.4 &44.2  \\ 
    RE-DETR&65.8  &68.6&64.7&63.3 &48.4 \\ 
\bottomrule

\end{tabular}
\vspace{0.5cm}
\caption{Result across unseen obstacles}
\label{zero-shot}
\end{table}

\subsection{Ablation Study}
To evaluate the effectiveness of the techniques employed in our data generation workflow, we conducted an ablation study. As shown in \cref{ablation}, each operation contributes differently to the overall performance. The Harmony module is the most effective, improving the mAP from 58.3\% to 62.1\%. In contrast, Random Texture has a negligible impact on performance. The combination of all three operations yields the highest overall accuracy, indicating that the techniques are complementary when used together.
\begin{table}[htb]
\centering
\begin{tabular}{c|cccc}
\toprule
 &Harmony&rescale&random-texture&mAP \\
\hline
 1&  &&   &58.3\\ 
  2& \checkmark &&&62.1\\ 
   3&  &\checkmark&&60.2   \\ 
   4&  &&\checkmark&   58.7\\ 
    5& \checkmark &\checkmark&\checkmark& 63.7  \\ 
\bottomrule
\end{tabular}
\vspace{0.5cm}
\caption{Results on ablation study }
\label{ablation}
\end{table}

\section{Discussion}
In this paper, we introduce SynRailObs, the first large-scale synthetic public synthetic dataset designed for railway intrusion scenarios. We leverage Segment Anything Model (SAM) and diffusion models to generate highly realistic representations of potential obstacles. The dataset is easily extensible by providing custom-designed prompts, without the need for manual annotations. Experimental results demonstrate the effectiveness of SynRailObs across varying distances, environments, and even in zero-shot settings. This dataset can be used to train obstacle detection models aimed at preventing accidents in railway environments. Nonetheless, SynRailObs has the following limitations:
\begin{itemize}
    \item There is an insufficient number of background images depicting extreme scenarios. Compared to normal environments, railway images captured under low illumination, rainstorms, or overexposure conditions are relatively rare on social media platforms.
    \item While image harmonization enhances the perceived realism of generated images, noticeable discrepancies persist when compared to real images, particularly with respect to shadows, lighting, and scale.
    \item  Currently, not all potential obstacle types are included in the dataset—an exhaustive inclusion is inherently infeasible. However, additional categories can be incorporated in future extensions.
\end{itemize}

\bibliographystyle{plain}
\bibliography{ref}
\newpage
\section*{NeurIPS Paper Checklist}

The checklist is designed to encourage best practices for responsible machine learning research, addressing issues of reproducibility, transparency, research ethics, and societal impact. Do not remove the checklist: {\bf The papers not including the checklist will be desk rejected.} The checklist should follow the references and follow the (optional) supplemental material.  The checklist does NOT count towards the page
limit. 

Please read the checklist guidelines carefully for information on how to answer these questions. For each question in the checklist:
\begin{itemize}
    \item You should answer \answerYes{}, \answerNo{}, or \answerNA{}.
    \item \answerNA{} means either that the question is Not Applicable for that particular paper or the relevant information is Not Available.
    \item Please provide a short (1–2 sentence) justification right after your answer (even for NA). 
\end{itemize}

{\bf The checklist answers are an integral part of your paper submission.} They are visible to the reviewers, area chairs, senior area chairs, and ethics reviewers. You will be asked to also include it (after eventual revisions) with the final version of your paper, and its final version will be published with the paper.

The reviewers of your paper will be asked to use the checklist as one of the factors in their evaluation. While "\answerYes{}" is generally preferable to "\answerNo{}", it is perfectly acceptable to answer "\answerNo{}" provided a proper justification is given (e.g., "error bars are not reported because it would be too computationally expensive" or "we were unable to find the license for the dataset we used"). In general, answering "\answerNo{}" or "\answerNA{}" is not grounds for rejection. While the questions are phrased in a binary way, we acknowledge that the true answer is often more nuanced, so please just use your best judgment and write a justification to elaborate. All supporting evidence can appear either in the main paper or the supplemental material, provided in appendix. If you answer \answerYes{} to a question, in the justification please point to the section(s) where related material for the question can be found.

IMPORTANT, please:
\begin{itemize}
    \item {\bf Delete this instruction block, but keep the section heading ``NeurIPS Paper Checklist"},
    \item  {\bf Keep the checklist subsection headings, questions/answers and guidelines below.}
    \item {\bf Do not modify the questions and only use the provided macros for your answers}.
\end{itemize}


\begin{enumerate}

\item {\bf Claims}
    \item[] Question: Do the main claims made in the abstract and introduction accurately reflect the paper's contributions and scope?
    \item[] Answer:\answerYes{} 
    \item[] Justification: Abstract states the contributions and scope

\item {\bf Limitations}
    \item[] Question: Does the paper discuss the limitations of the work performed by the authors?
    \item[] Answer: \answerYes{} 
    \item[] Justification: in discussion section, we discuss the limitations of
    the work

\item {\bf Theory assumptions and proofs}
    \item[] Question: For each theoretical result, does the paper provide the full set of assumptions and a complete (and correct) proof?
    \item[] Answer: \answerNA{}
    \item[] Justification: No complete proof provided

    \item {\bf Experimental result reproducibility}
    \item[] Question: Does the paper fully disclose all the information needed to reproduce the main experimental results of the paper to the extent that it affects the main claims and/or conclusions of the paper (regardless of whether the code and data are provided or not)?
    \item[] Answer: \answerYes{} 
    \item[] Justification: \justificationTODO{}

\item {\bf Open access to data and code}
    \item[] Question: Does the paper provide open access to the data and code, with sufficient instructions to faithfully reproduce the main experimental results, as described in supplemental material?
    \item[] Answer: \answerYes{} 
    \item[] Justification: dataset and code are all avaliable

\item {\bf Experimental setting/details}
    \item[] Question: Does the paper specify all the training and test details (e.g., data splits, hyperparameters, how they were chosen, type of optimizer, etc.) necessary to understand the results?
    \item[] Answer:\answerYes{}
    \item[] Justification: yes, in experiment setting part

\item {\bf Experiment statistical significance}
    \item[] Question: Does the paper report error bars suitably and correctly defined or other appropriate information about the statistical significance of the experiments?
    \item[] Answer: \answerTODO{} 
    \item[] Justification: \justificationTODO{}

\item {\bf Experiments compute resources}
    \item[] Question: For each experiment, does the paper provide sufficient information on the computer resources (type of compute workers, memory, time of execution) needed to reproduce the experiments?
    \item[] Answer: \answerYes{} 
    \item[] Justification: yes, in setting part
    
\item {\bf Code of ethics}
    \item[] Question: Does the research conducted in the paper conform, in every respect, with the NeurIPS Code of Ethics \url{https://neurips.cc/public/EthicsGuidelines}?
    \item[] Answer: \answerYes{} 
    \item[] Justification: \justificationTODO{}

\item {\bf Broader impacts}
    \item[] Question: Does the paper discuss both potential positive societal impacts and negative societal impacts of the work performed?
    \item[] Answer: \answerYes{} 
    \item[] Justification: in discussion part
    
\item {\bf Safeguards}
    \item[] Question: Does the paper describe safeguards that have been put in place for responsible release of data or models that have a high risk for misuse (e.g., pretrained language models, image generators, or scraped datasets)?
    \item[] Answer: \answerYes{} 
    \item[] Justification: in discussion part

\item {\bf Licenses for existing assets}
    \item[] Question: Are the creators or original owners of assets (e.g., code, data, models), used in the paper, properly credited and are the license and terms of use explicitly mentioned and properly respected?
    \item[] Answer:\answerYes{}
    \item[] Justification: we introduce all tools and datasets we used

\item {\bf New assets}
    \item[] Question: Are new assets introduced in the paper well documented and is the documentation provided alongside the assets?
    \item[] Answer: \answerYes{} 
    \item[] Justification: \justificationTODO{}

\item {\bf Crowdsourcing and research with human subjects}
    \item[] Question: For crowdsourcing experiments and research with human subjects, does the paper include the full text of instructions given to participants and screenshots, if applicable, as well as details about compensation (if any)? 
    \item[] Answer: \answerNo{} 
    \item[] Justification: used previous dataset of human subjects

\item {\bf Institutional review board (IRB) approvals or equivalent for research with human subjects}
    \item[] Question: Does the paper describe potential risks incurred by study participants, whether such risks were disclosed to the subjects, and whether Institutional Review Board (IRB) approvals (or an equivalent approval/review based on the requirements of your country or institution) were obtained?
    \item[] Answer: \answerNo{} 
    \item[] Justification: No related risk

\item {\bf Declaration of LLM usage}
    \item[] Question: Does the paper describe the usage of LLMs if it is an important, original, or non-standard component of the core methods in this research? Note that if the LLM is used only for writing, editing, or formatting purposes and does not impact the core methodology, scientific rigorousness, or originality of the research, declaration is not required.
    \item[] Answer: \answerNA{} 
    \item[] Justification: no LLM used in experiments

\end{enumerate}
\end{document}